\documentclass{sig-alternate-2013}

\usepackage{tabularx} 
\usepackage{amssymb}
\usepackage{graphicx}
\usepackage{url}   
\usepackage{booktabs}
\usepackage{textcomp}
\usepackage{epstopdf}
\usepackage[
  pass,% keep layout unchanged 
]{geometry}

\newcommand{\eg}{e.\,g., }
\newcommand{\ie}{i.\,e., }

\newfont{\mycrnotice}{ptmr8t at 7pt}
\newfont{\myconfname}{ptmri8t at 7pt}
\let\crnotice\mycrnotice%
\let\confname\myconfname%

\permission{Permission to make digital or hard copies of all or part of this work for personal or classroom use is granted without fee provided that copies are not made or distributed for profit or commercial advantage and that copies bear this notice and the full citation on the first page. Copyrights for components of this work owned by others than the author(s) must be honored. Abstracting with credit is permitted. To copy otherwise, or republish, to post on servers or to redistribute to lists, requires prior specific permission and/or a fee. Request permissions from Permissions@acm.org.}
\conferenceinfo{AVEC'16,}{October 16, 2016, Amsterdam, NL} 
\copyrightetc{Copyright is held by the owner/author(s). Publication rights licensed to ACM.\\ACM \the\acmcopyr}
\crdata{978-1-4503-4516-3/16/10\ldots \$15.00.\\
DOI: http://dx.doi.org/10.1145/2988257.2988258}
\begin{document}

%
% --- Author Metadata here ---
\conferenceinfo{AVEC'16}{16 October 2016, Amsterdam, NL}
%\CopyrightYear{2007} % Allows default copyright year (20XX) to be over-ridden - IF NEED BE.
%\crdata{0-12345-67-8/90/01}  % Allows default copyright data (0-89791-88-6/97/05) to be over-ridden - IF NEED BE.
% --- End of Author Metadata ---

\title{AVEC 2016 --  Depression, Mood, and Emotion Recognition Workshop and Challenge}

\numberofauthors{6} %  in this sample file, there are a *total*
\author{
% 1st. author
\alignauthor
Michel Valstar\\
       \affaddr{University of Nottingham}\\
       \affaddr{School of Computer Science}
% 2nd. author
\alignauthor
Jonathan Gratch\\
      \affaddr{University of Southern California}\\
       \affaddr{ICT}
% 3rd. author  
\alignauthor 
Bj\"{o}rn Schuller\titlenote{The author is further affiliated with Imperial College London, Department of Computing, London, U.K.}\\
       \affaddr{University of Passau}\\
        \affaddr{Chair of Complex \& Intelligent Systems}
\and  % use '\and' if you need 'another row' of author names
% 4th. author
\alignauthor 
Fabien Ringeval\titlenote{The author is further affiliated with University of Passau, Chair of Complex \& Intelligent Systems}\\
      \affaddr{Universit\'e Grenoble Alpes}\\
       \affaddr{Laboratoire d'Informatique de Grenoble}
% 5th. author
\alignauthor 
Denis Lalanne\\
       \affaddr{University of Fribourg}\\
       \affaddr{Human-IST Research Center}\\
% 6th. author
\alignauthor
Mercedes Torres Torres\\
       \affaddr{University of Nottingham}\\
       \affaddr{School of Computer Science}
\and  % use '\and' if you need 'another row' of author names
% 7th. author
\alignauthor 
Stefan Scherer\\
      \affaddr{University of Southern California}\\
       \affaddr{ICT}
% 8th. author
\alignauthor Giota Stratou\\
      \affaddr{University of Southern California}\\
       \affaddr{ICT}
% 9th. author
\alignauthor Roddy Cowie\\
      \affaddr{Queen's University Belfast}\\
       \affaddr{Department of Psychology}     
\and  % use '\and' if you need 'another row' of author names
% 10th. author
\alignauthor Maja Pantic\titlenote{The author is further affiliated with Twente University, EEMCS, Twente, The Netherlands.}\\
       \affaddr{Imperial College London}\\
       \affaddr{Intelligent Behaviour Understanding Group}
}
% There's nothing stopping you putting the seventh, eighth, etc.
% author on the opening page (as the 'third row') but we ask,
% for aesthetic reasons that you place these 'additional authors'
% in the \additional authors block, viz.
%\additionalauthors{Additional authors: John Smith (The Th{\o}rv{\"a}ld Group,
%email: {\texttt{jsmith@affiliation.org}}) and Julius P.~Kumquat
%(The Kumquat Consortium, email: {\texttt{jpkumquat@consortium.net}}).}
\date{25 April 2016}
% Just remember to make sure that the TOTAL number of authors
% is the number that will appear on the first page PLUS the
% number that will appear in the \additionalauthors section.

\maketitle
\begin{abstract}
 
The Audio/Visual Emotion Challenge and Workshop (AVEC 2016) ``Depression, Mood and Emotion'' will be the sixth competition event aimed at comparison of multimedia processing and machine learning methods for automatic audio, visual and physiological depression and emotion analysis, with all participants competing under strictly the same conditions. The goal of the Challenge is to provide a common benchmark test set for multi-modal information processing and to bring together the depression and emotion recognition communities, as well as the audio, video and physiological processing communities, to compare the relative merits of the various approaches to depression and emotion recognition under well-defined and strictly comparable conditions and establish to what extent fusion of the approaches is possible and beneficial. This paper presents the challenge guidelines, the common data used, and the performance of the baseline system on the two tasks.

\end{abstract}

%\category{J}{Computer Applications}{Miscellaneous}
%A category including the fourth, optional field follows...
%\category{D.2.8}{Software Engineering}{Metrics}[complexity measures, performance measures]



\section{Introduction}
\noindent The 2016 Audio-Visual Emotion Challenge and Workshop (AVEC 2016) will be the sixth competition event aimed at comparison of multimedia 
%FR: changed here to include physiology and depression
processing and machine learning methods for automatic audio, video, and physiological analysis of emotion and depression, with all participants competing under strictly the same conditions. 
%FR: same for physiology here
The goal of the Challenge is to compare the relative merits of the approaches (audio, video, and/or physiologic) to emotion recognition and severity of depression estimation under well-defined and strictly comparable conditions, and establish to what extent fusion of the approaches is possible and beneficial. A second motivation is the need to advance emotion recognition for multimedia retrieval to a level where behaviomedical systems \cite{Valstar2014_ABU} are able to deal with large volumes of non-prototypical naturalistic behaviour in reaction to known stimuli, as this is exactly the type of data that diagnostic and in particular monitoring tools, as well as other applications, would have to face in the real world.

%FR: added missing AVEC 2014 paper in the bib file
AVEC 2016 will address emotion and depression recognition. The emotion recognition sub-challenge is a refined re-run of the AVEC 2015 challenge \cite{RingevalEtAl2015_FAR}, largely based on the same dataset. The depression severity estimation sub-challenge is based on a novel dataset of human-agent interactions, and sees the return of depression analysis, which was a huge success in the AVEC 2013 \cite{ValstarEtAl13_ACA} and 2014 \cite{Valstar14-A2T} challenges.

\begin{itemize}
	\item \textbf{Depression Classification Sub-Challenge} (DCC): participants are required to classify whether a person is classified as depressed or not, where the binary ground-truth is based on the severity of self-reported depression as indicated by the PHQ-8 score for every human-agent interaction. For the DCC, performance in the competition will be measured using the average \textbf{F1 score} for both classes \textit{depressed} and \textit{not\_depressed}. Participants are encouraged to provide an estimate of the severity of depression, by calculating the root mean square error over all HCI experiment sessions between the predicted and ground-truth PHQ-8 score. In addition, participants are also encouraged to report on overall accuracy, average precision, and average recall to further analyse their results in the paper accompanying their submission.
    
	\item \textbf{Multimodal Affect Recognition Sub-Challenge} (MASC) participants are required to perform fully continuous affect recognition of two affective dimensions: Arousal, and Valence, where the level of affect has to be predicted for every moment of the recording. For the MASC, two regression problems need to be solved: prediction of the continuous dimensions  \textsc{Valence} and \textsc{Arousal}. The MASC competition measure is the \textbf{Concordance Correlation Coefficient (CCC)}, which combines the Pearson's correlation coefficient (CC) with the square difference between the mean of the two compared time series, as shown in \ref{eq:ccc}.
\begin{equation}\label{eq:ccc}
\rho_c=\frac{2\rho\sigma_x\sigma_y}{\sigma_x^2+\sigma_y^2+(\mu_x-\mu_y)^2}
\end{equation} 
where $\rho$ is the Pearson correlation coefficient between two time series
(\eg prediction and gold-standard), $\sigma_{x}^2$ and $\sigma_{y}^2$ is the variance of each time series, and $\mu_x$ and $\mu_y$ are the mean value of each. Therefore, predictions that are well correlated with the gold standard but shifted in value are penalised in proportion to the deviation.

\end{itemize}

%Both Sub-Challenges allow contributors to find their own features to use with their regression algorithm. In addition, standard feature sets are provided (for audio and video separately), which participants are free to use. The labels of the test partition remain unknown to the participants, and participants have to stick to the definition of training, development, and test partition. They may freely report on results obtained on the development partition, but are limited to five trials per Sub-Challenge in submitting their results on the test partition. 

%FR: "Only contributions with a relevant accepted paper will be eligible for challenge participation" -> there was some discussions on this last year, and at the end, all entries were ranked, without visual distinctions (e.g., gray shaded bars) in the plot recapitulating the results between accepted and rejected papers, or team that did not submit a paper. Do we want to go back on this and make visual distinctions, or do we keep ranking all submissions (preferred)?
To be eligible to participate in the challenge, every entry has to be accompanied by a paper presenting the results and the methods that created them, which will undergo peer-review. Only contributions with a relevant accepted paper will be eligible for challenge participation. The organisers reserve the right to re-evaluate the findings, but will not participate in the Challenge themselves. 

\section{Depression Analysis Corpus}\label{s:depressionDatabase}

\noindent The Distress Analysis Interview Corpus - Wizard of Oz (DAIC-WOZ) database is part of a larger corpus, the Distress Analysis Interview Corpus (DAIC) \cite{GratchEtAl2014_DAI}, that contains clinical interviews designed to support the diagnosis of psychological distress conditions such as anxiety, depression, and post-traumatic stress disorder. These interviews were collected as part of a larger effort to create a computer agent that interviews people and identifies verbal and nonverbal indicators of mental illness \cite{DeVaultEtAl2014_SVH}. Data collected include audio and video recordings and extensive questionnaire responses; this part of the corpus includes the Wizard-of-Oz interviews, conducted by an animated virtual interviewer called Ellie, controlled by a human interviewer in another room. Data has been transcribed and annotated for a variety of verbal and non-verbal features.

Information on how to obtain shared data can be found in this location: http://dcapswoz.ict.usc.edu. Data is freely available for research purposes.

\subsection{Depression Analysis Labels}\label{s:depressionLabels}

\begin{figure}[!t]
\begin{center}
\includegraphics[width=0.8\columnwidth]{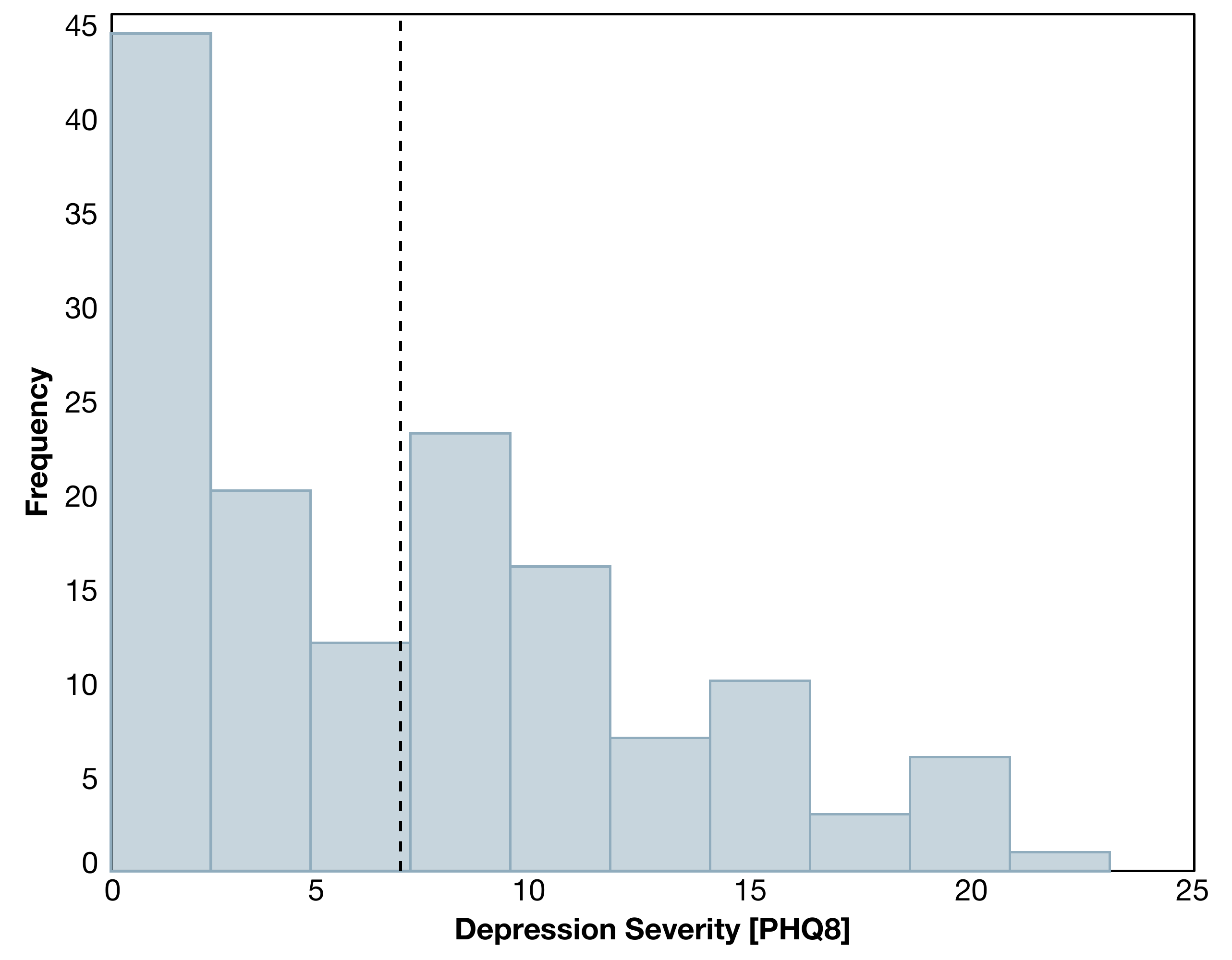}
\end{center}
\caption{{\bf Histogram of depression severity scores for DESC challenge. Data of training and development set are provided here.}}
\label{fig:DESC_hist}
\end{figure}

\noindent The level of depression is labelled with a single value per recording using a standardised self-assessed subjective depression questionnaire, the PHQ-8 \cite{KroenkeEtAl2009_PMC}. This is similar to the PHQ-9 questionnaire, but with the suicidal ideation question removed for ethical reasons. The average depression severity on the training and development set of the challenge is $M = 6.67$ ($SD = 5.75$). The distribution of the depression severity scores based on the challenge training and development set is provided in Figure \ref{fig:DESC_hist}. A baseline classifier that constantly predicts the mean score of depression provides an $RMSE = 5.73$ and an $MAE = 4.74$.

\subsection{Depression Analysis Baseline Features}\label{s:features}

\noindent In the following sections we describe how the publicly available baseline feature sets are computed for either the audio or the video data. Participants can use these feature sets exclusively or in addition to their own features. For ethical reasons, no raw video is made available.

\subsubsection{Video Features}
Based on the \textit{OpenFace}~\cite{baltruvsaitisopenface} framework\footnote{\url{https://github.com/TadasBaltrusaitis/CLM-framework}}, we provide different types of video features: 
\begin{itemize}
\item facial landmarks: 2D and 3D coordinates of 68 points on the face, estimated from video
\item HOG (histogram of oriented gradients) features on the aligned 112x112 area of the face
\item gaze direction estimate for both eyes
\item head pose: 3D position and orientation of the head
\end{itemize}

In addition to that, we provide emotion and facial action unit continuous measures based on \textit{FACET} software\cite{littlewort2011computer}. Specifically, we provide the following measures:
\begin{itemize}
\item emotion: \{Anger, Contempt, Disgust, Joy, Fear, Neutral, Sadness, Surprise, Confusion, Frustration\}
\item AUs: \{AU1, AU2, AU4, AU5, AU6, AU7, AU9, AU10, AU12, AU14, AU15, AU17, AU18, AU20, AU23, AU24, AU25, AU26, AU28, AU43\}
\end{itemize}

\subsubsection{Audio Features}

For the audio features we utilized COVAREP(v1.3.2), a freely available open source Matlab and Octave toolbox for speech analyses \cite{degottex2014covarep}\footnote{\url{http://covarep.github.io/covarep/}}. The toolbox comprises well validated and tested feature extraction methods that aim to capture both voice quality as well as prosodic characteristics of the speaker. These methods have been successfully shown to be correlated with psychological distress and depression in particular \cite{scherer2014automatic,Scherer_etAl2015}. In particular, we extracted the following features:

\begin{itemize}
\item \textbf{Prosodic: }Fundamental frequency (F0) and voicing (VUV)
\item \textbf{Voice Quality: }Normalized amplitude quotient (NAQ), Quasi open quotient (QOQ), the difference in amplitude of the first two harmonics of the differentiated glottal source spectrum (H1H2), parabolic spectral parameter (PSP), maxima dispersion quotient (MDQ), spectral tilt/slope of wavelet responses (peakSlope), and shape parameter of the Liljencrants-Fant model of the glottal pulse dynamics (Rd)
\item \textbf{Spectral:} Mel cepstral coefficients (MCEP0-24), Harmonic Model and Phase Distortion mean (HMPDM0-24) and deviations (HMPDD0-12).
\end{itemize}

\noindent In addition to the feature set above, raw audio and transcripts of the interview are being provided, allowing the participants to compute additional features on their own. For more details on the shared features and the format of the files participants should also review the DAIC-WOZ documentation \footnote{\url{http://dcapswoz.ict.usc.edu/wwwutil_files/DAICWOZDepression_Documentation.pdf}}.

\section{Emotion Analysis Corpus}

The Remote Collaborative and Affective Interactions (RECOLA) database \cite{Ringeval13-ITR} was recorded to study socio-affective behaviours from multimodal data in the context of computer supported collaborative work \cite{Ringeval13-OTI}. Spontaneous and naturalistic interactions were collected during the resolution of a collaborative task that was performed in dyads and remotely through video conference. Multimodal signals, \ie audio, video, electro-cardiogram (ECG) and electro-dermal activity (EDA), were synchronously recorded from 27 French-speaking subjects. Even though all subjects speak French fluently, they have different nationalities (i. e., French, Italian or German), which thus provide some diversity in the expression of emotion.

Data is freely available for research purposes, information on how to obtain the RECOLA database can be found on this location: \url{http://diuf.unifr.ch/diva/recola}.

\subsection{Emotion Analysis Labels}

Regarding the annotation of the dataset, time-continuous ratings (40 ms binned frames) of emotional arousal and valence were created by six gender balanced French-speaking assistants for the first five minutes of all recordings, because participants discussed more about their strategy -- hence showing emotions -- at the beginning of their interaction.

To assess inter-rater reliability, we computed the intra-class correlation coefficient (ICC(3,1)) \cite{Shrout79-ICU}, and Cronbach's $\alpha$~\cite{Cronbach51-CAA}; ratings are concatenated over all subjects. 
Additionally, we computed the root-mean-square error (RMSE), Pearson's CC and the CCC \cite{Li89-ACC}; values are averaged over the $C_2^6$ pairs of raters. 
Results indicate a very strong inter-rater reliability for both arousal and valence, cf. Table \ref{tab:ira}. 
A normalisation technique based on the Evaluator Weighted Estimator~\cite{Grimm05-EON}, is used prior to the computation of the gold-standard, \ie the average of all ratings for each subject~\cite{Ringeval15-POA}.
This technique has significantly ($p<0.001$ for CC) improved the inter-rater reliability for both arousal and valence; the Fisher Z-transform is used to perform statistical comparisons between CC in this study.

The dataset was divided into speaker disjoint subsets for training, development (validation) and testing, by stratifying (balancing) on gender and mother tongue, cf.\ Table \ref{tab:recola_strat}.

\begin{table}[t]
\centering
\caption{Inter-rater reliability on arousal and valence for the 6 raters and the 27 subjects of the RECOLA database; raw or normalised ratings \cite{Ringeval15-POA}.}
\label{tab:ira}
\begin{tabular}{c|c|c|c|c|c} \midrule
&RMSE &CC & CCC &ICC &$\alpha$\\\midrule
\multicolumn{6}{c}{\textit{Raw}}\\\midrule
Arousal &.344 &.400 &.277 &.775 &.800\\ 
Valence &.218 &.446 &.370 &.811 &.802\\\midrule
\multicolumn{6}{c}{\textit{Normalised}}\\\midrule
Arousal &.263 &.496 &.431 &.827 &.856\\ 
Valence &.174 &.492 &.478 &.844 &.829\\ 
\end{tabular}
\end{table}

\begin{table}[!t]
\caption{Partitioning of the \textsc{RECOLA} database into train, development, and test sets.}
\label{tab:recola_strat}
\begin{center}
\begin{tabular}{l|r|r|r}\midrule
$\#$ 	 				& train		& dev		& test\\\midrule
female				& 6			& 5			& 5\\
male					& 3			& 4			& 4\\\midrule
French				& 6			& 7			& 7\\
Italian				& 2			& 1			& 2\\
German				& 1			& 1			& 0\\\midrule
age $\mu$ ($\sigma$)	& 21.2 (1.9) 	& 21.8 (2.5)	& 21.2 (1.9)\\
\end{tabular}
\end{center}
\end{table}

\subsection{Emotion Analysis Baseline Features}

In the followings we describe how the baseline feature sets are computed for video, audio, and physiological data.

\subsubsection{Video Features}
Facial expressions play an important role in the communication of emotion~\cite{Ekman02-FAC}. 
Features are usually grouped in two types of facial descriptors: appearance and geometric based \cite{ValstarEtAl2015_FER}. 
For the video baseline features set, we computed both, using Local Gabor Binary Patterns from Three Orthogonal Planes (LGBP-TOP)~\cite{Almaev13-LGB} for appearance and facial landmarks~\cite{Xiong13-SDM} for geometric.

The LGBP-TOP are computed by splitting the video into spatio-temporal video volumes. 
Each slice of the video volume extracted along 3 orthogonal planes ($x$-$y$, $x$-$t$ and $y$-$t$) is first convolved with a bank of 2D Gabor filters. 
The resulting Gabor pictures in the direction of $x$-$y$ plane are divided into 4x4 blocks. 
In the $x$-$t$ and $y$-$t$ directions they are divided into 4x1 blocks. 
The LBP operator is then applied to each of these resulting blocks followed by the concatenation of the resulting LBP histograms from all the blocks. 
A feature reduction is then performed by applying a Principal Component Analysis (PCA) from a low-rank (up to rank 500) approximation~\cite{Halko11-AAF}.
We obtained 84 features representing 98\,\% of the variance. 

In order to extract geometric features, we tracked 49 facial landmarks with the Supervised Descent Method (SDM)~\cite{Xiong13-SDM} and aligned them with a mean shape from stable points (located on the eye corners and on the nose region). 
As features, we computed the difference between the coordinates of the aligned landmarks and those from the mean shape, and also between the aligned landmark locations in the previous and the current frame; this procedure provided 196 features in total. 
We then split the facial landmarks into groups according to three different regions: i) the left eye and left eyebrow, ii) the right eye and right eyebrow and iii) the mouth. 
For each of these groups, the Euclidean distances (L2-norm) and the angles (in radians) between the points are computed, providing 71 features. 
We also computed the Euclidean distance between the median of the stable landmarks and each aligned landmark in a video frame. 
In total the geometric set includes 316 features. 

Both appearance and geometric feature sets are interpolated by a piecewise cubic Hermite polynomial to cope with dropped frames.
Finally, the arithmetic mean and the standard-deviation are computed on all features using a sliding window, which is shifted forward at a rate of 40\,ms.

\subsubsection{Audio Features}
In contrast to large scale feature sets, which have been successfully applied to many speech classification tasks~\cite{Schuller13-TI2, Schuller14-TI2}, smaller, expert-knowledge based feature sets have also shown high robustness for the modelling of emotion from speech~\cite{Ringeval14-ERI,Bone14-RUA}.
Some recommendations for the definition of a minimalistic acoustic standard parameter set have been recently investigated, 
and have led to the Geneva Minimalistic Acoustic Parameter Set (\textsc{GeMAPS}), and to an extended version (\textsc{eGeMAPS})~\cite{Eyben15-TGM}, which is used here as baseline. 
The acoustic low-level descriptors (LLD) cover spectral, cepstral, prosodic and voice quality information and are extracted with the \textsc{openSMILE} toolkit~\cite{Eyben13-RDI}.%, cf. Table \ref{tab:gemaps_lld}. 

%\begin{table}[t]
%\caption{32 acoustic low-level descriptors (LLDs); $^1$computed on voiced and unvoiced frames; $^2$computed on voiced and all frames; $^3$computed on voiced, unvoiced, and all frames.}
%\centering
%\begin{tabular}{l|l}
%\hline
%\textbf{1 energy-related LLD}  & \textbf{Group}\\\midrule
%Sum of auditory spectrum (loudness) &  Prosodic \\\midrule%1 LLD
%\textbf{19 spectral-related LLDs} & \textbf{Group} \\\midrule
%$\alpha$ ratio (50--1000\,Hz / 1--5\,kHz)$^1$ & Spectral\\ %4 LLD
%Energy slope (0--500\,Hz, 0.5--1.5\,kHz)$^1$ & Spectral\\ %4 LLD
%Hammarberg index$^1$ & Spectral\\ %2 LLD
%MFCC 1--4$^2$ & Cepstral\\ %12 LLD
%Spectral flux$^3$ & Spectral\\\midrule %3 LLD
%\textbf{12 voicing-related LLDs} & \textbf{Group} \\\midrule
%$F_0$ (semi-tone) & Prosodic \\ %2 LLD
%Formants 1, 2, 3 (freq., bandwidth, ampl.) & Voice qual. \\ %9 LLD
%Harmonic difference H1--H2, H1--A3 & Voice qual. \\ %2 LLD
%Log.\ HNR, jitter (local), shimmer (local) & Voice qual. \\ %3 LLD
%\end{tabular}
%\label{tab:gemaps_lld}
%\end{table}

As the data in the RECOLA database contains long continuous recordings, we used overlapping fixed length segments, which are shifted forward at a rate of 40\,ms, to extract functionals; the arithmetic mean and the coefficient of variation are computed on all 42 LLD.
To pitch and loudness the following functionals are additionally applied: percentiles 20, 50 and 80, the range of percentiles 20 -- 80 and the mean and standard deviation of the slope of rising/falling signal parts.
Functionals applied to the pitch, jitter, shimmer, and all formant related LLDs, are applied to voiced regions only.
Additionally, the average RMS energy is computed and 6 temporal features are included: the rate of loudness peaks per second, mean length and standard deviation of continuous voiced and unvoiced segments and the rate of voiced segments per second, approximating the pseudo syllable rate.
Overall, the acoustic baseline features set contains 88 features.

\subsubsection{Physiological Features}

Physiological signals are known to be well correlated with emotion~\cite{Koelstra12-DAD,Knapp11-PSA}, despite not being directly perceptible the way audio-visual are. 
Although there are some controversies about peripheral physiology and emotion~\cite{Schachter12-CAP,Keltner10-E}, we believe that autonomic measures should be considered along with audio-visual data in the realm of affective computing, as they do not only provide complementary descriptions of affect, but can also be easily and continuously monitored with wearable sensors~\cite{Sano14-QAO,Picard14-AMA,Chen15-AAI}.

\begin{table*}
\begin{center}
\caption{Baseline results for depression classification. Performance is measured in F1 score for \emph{depressed} and \emph{not depressed} classes as reported through the PHQ-8. In addition, precision and recall are provided. Values for class \emph{not depressed} are reported in brackets.}
\vspace{2mm}
\label{t:baseline_DCC}
\begin{tabular}{l | l | c|c|c}
\midrule
Partition & Modality & F1 score & Precision & Recall \\\midrule
Development & Audio&  .462 (.682) & .316 (.938) & .857 (0.54) \\ 
Development & Video &  .500 (.896) & .600 (.867) & .428 (.928) \\
Development & Ensemble &  .500 (.896) & .600 (.867) & .428 (.928) \\ \midrule
Test & Audio & .410 (.582) & .267 (.941) & .889 (.421) \\
Test & Video &  .583 (.851)  & .467 (.938)  & .778 (.790) \\
Test & Ensemble &  .583 (.857)  & .467 (.938)  & .778 (.790) \\ 
\end{tabular}
\end{center}
\end{table*}

As baseline features, we extracted features from both ECG and EDA signals with overlapping (step of 40\,ms) windows.
The ECG signal was firstly band-pass filtered ($[3-27]$ Hz) with a zero-delay 6th order Butterworth filter \cite{Ringeval15-POA}, and 19 features were then computed: the zero-crossing rate, the four first statistical moments, the normalised length density, the non-stationary index, the spectral entropy, slope, mean frequency plus 6 spectral coefficients, the power in low frequency (LF, 0.04-0.15\,Hz), high frequency (HF, 0.15-0.4\,Hz) and the LF/HF power ratio.
Additionally, we extracted the heart rate (HR) and its measure of variability (HRV) from the filtered ECG signal \cite{Ringeval15-POA}. For each of those two descriptors, we computed the two first statistical moments, the arithmetic mean of rising and falling slope, and the percentage of rising values, which provided 10 features in total.

EDA reflects a rapid, transient response called skin conductance response (SCR), as well as a slower, basal drift called skin conductance level (SCL) \cite{Dawson00-TES}. Both, SCL (0--0.5\,Hz) and SCR (0.5--1\,Hz) are estimated using a 3rd order Butterworth filter, 8 features are then computed for each of those three low-level descriptors: the four first statistical moments from the original time-series and its first order derivate w.r.t. time.

\section{Challenge Baselines}\label{s:baseline}

\noindent For transparency and reproducibility, we use standard and open-source algorithms for both sub-challenges. We describe below how the baseline system was defined and the results we obtained for each modality separately, as well as on the fusion of all modalities.
%For both sub-challenges, we conducted three separate baselines: one using video features only, another one using audio features only and finally an audio-visual baseline that simply combines the results of the previous two. 

\subsection{Depression}

The challenge baseline for the depression classification sub-challenge is computed using the scikit-learn toolbox\footnote{\url{http://scikit-learn.org/}}. In particular, we fit a linear support vector machine with stochastic gradient descent, i.\,e. the loss is computed one sample at a time and the model is sequentially updated. We validated the model on the development set and conducted a grid search for optimal hyper-parameters on the development set of both the audio data and video data separately. Features of both modalities are taken from the provided challenge baseline features. Classification and training was performed on a frame-wise basis (i.e., at 100Hz for audio and 30Hz for video); temporal fusion was conducted through simple majority voting of all the frames within an entire screening interview. For both modalities we conducted a grid search for the following parameters: loss function $\in \{\mbox{logarithmic}, \mbox{hinge loss}\}$,  regularization $\in \{\mbox{L1}, \mbox{L2}\}$, and $\alpha \in \{1e1, 1e0, \dots, 1e-5\}$. For the audio data the optimal identified hyper-parameters are loss function $= {\mbox{hinge loss}}$,  regularization $= {\mbox{L1}}$, and $\alpha = {1e-3}$. For the video data the optimal identified hyper-parameters are loss function $= {\mbox{logarithmic}}$,  regularization $= {\mbox{L1}}$, and $\alpha = {1e0}$. The ensemble of audio and video was computed through a simple binary fusion of a logical AND. The test performance was computed on a classifier trained using the found optimal parameters from the grid search. Since the positive outputs of the video modality are a subset of those of the audio the ensemble classifier's performance is exactly the same as the video modality for both the development and test sets. Results are summarized in Table \ref{t:baseline_DCC}.

\begin{table}
\begin{center}
\caption{Baseline results for depression severity estimation. Performance is measured in mean absolute error (MAE) and root mean square error (RMSE) between the predicted and reported PHQ-8 scores, averaged over all sequences.}
\vspace{2mm}
\label{t:baseline_DSC}
\begin{tabular}{l | l | c|c }
\midrule
%Partition & Modality & RMSE & MAE \\\midrule
%Development & Audio & 6.7418 &  5.3566 \\
%Development & Video &  7.1332  & 5.8767 \\
%Development & Audio-Video & 6.6212 & 5.5222 \\\midrule
%Test & Audio&  7.7758 & 5.7224 \\
%Test & Video &  6.9692 & 6.1154  \\
%Test & Audio-Video &  7.0467 & 5.6567 \\
Partition & Modality & RMSE & MAE \\\midrule
Development & Audio & 6.74 &  5.36 \\
Development & Video &  7.13  & 5.88 \\
Development & Audio-Video & 6.62 & 5.52 \\\midrule
Test & Audio&  7.78 & 5.72 \\
Test & Video &  6.97 & 6.12  \\
Test & Audio-Video &  7.05 & 5.66 \\
\end{tabular}
\end{center}
\end{table}

In addition to classification baseline, we also computed a regression baseline using random forest regressor. The only hyper-parameter in this experiment was the number of trees $\in {10, 20, 50, 100, 200}$ in the random forest. For both audio and video the best performing random forest has trees $= 10$. Regression was performed on a frame-wise basis as the classification and temporal fusion over the interview was conduced by averaging of outputs over the entire screening interview. Fusion of audio and video modalities was performed by averaging the regression outputs of the unimodal random forest regressors. The performance for both root mean square error (RMSE) and mean absolute error (MAE) for development and test sets is provided in Table \ref{t:baseline_DSC}.

\subsection{Affect}

\begin{table}[t]
\caption{Size of the window $W$ in seconds used to extract features on the different modalities, and delay $D$ in seconds applied to the gold-standard, according to the emotional dimension, \ie arousal ($A$), and valence ($V$); parameters were obtained as the result of an optimisation of the performance measured as CCC on the development partition.}
\centering
\begin{tabular}{l|l|l|l|l}\midrule
&\multicolumn{2}{c|}{Arousal} &\multicolumn{2}{c}{Valence} \\\midrule%
Modality &$W_A$ &$D_A$ &$W_V$ &$D_V$ \\\midrule
Audio &4 &2.8 &6 &3.6 \\ 
Video-appearance &6 &2.8 &4 &2.4 \\ 
Video-geometric &4 &2.4 &8 &2.8 \\ 
ECG &4 &0.4 &10 &2.0 \\ 
HRHRV &8 &0.0 &8 &0.0 \\ 
EDA &8 &0.0 &10 &0.4 \\ 
SCL &4 &0.0 &14 &2.4 \\ 
SCR &4 &0.8 &14 &0.8 \\ 
\end{tabular}
\label{tab:wsize}
\end{table}

\begin{table}[t!]
\caption{Baseline results for affect recognition on the development (D) and test (T) partitions from audio, video (appearance and geometric), and physiologic (ECG, HRHRV, EDA, SCL, and SCR) feature sets, and their late fusion (multimodal). Performance is measured in Concordance correlation coefficient.}% and for raw predictions or
\centering
\begin{tabular}{l|c|c}
\midrule
Modality &Arousal &Valence\\ \midrule
D-Audio &.796 &.455 \\ 
D-Video-appearance &.483 &.474 \\
D-Video-geometric &.379 &.612\\ 
D-ECG &.271 &.153\\
D-HRHRV &.379 &.293\\
D-EDA &.073 &.194 \\
D-SCL &.068 &.166 \\
D-SCR &.073 &.085\\\midrule
D-Multimodal &\textbf{.821} &\textbf{.683}\\\midrule
T-Audio &.648 &.375 \\ 
T-Video-appearance &.343 &.486 \\
T-Video-geometric &.272 &.507\\ 
T-ECG &.158 &.121\\
T-HRHRV &.334 &.198\\
T-EDA &.075 &.228 \\
T-SCL &.066 &.216 \\
T-SCR &.065 &.145\\\midrule
T-Multimodal &\textbf{.683} &\textbf{.639}\\
\end{tabular}
\label{t:baseline_MASC}
\end{table}

\begin{figure*}
\centering
\includegraphics[width=1.6\columnwidth]{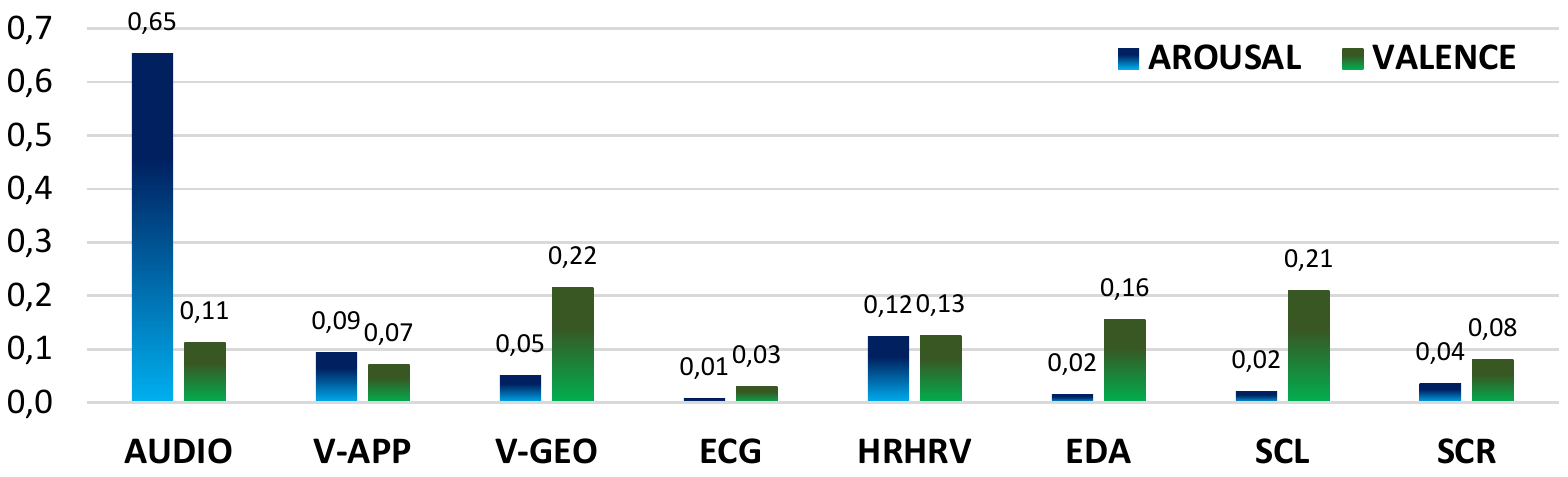}
\caption{\label{f:MASC_fusion}Percentage of contribution of each modality in the prediction of emotion; values are derived from the multimodal fusion model; V-APP: video appearance; V-GEO: video geometric; ECG: electrocardiogram; HRHV: heart rate and heart rate variability; EDA: electrodermal activity; SCL: skin conductance level; SCR: skin conductance resistance.}
\end{figure*}

Mono-modal emotion recognition was first investigated separately for each modality. 
Baseline features were extracted as previously described, with a window size $W$ ranging from four to 14 seconds, and a step of two seconds. The window was centred, \ie the first feature vector was assigned to the center of the window ($W/2$), and duplicated for the previous frames; the same procedure was applied for the last frames. For video data, frames for which the face was not detected were ignored. For EDA, SCL, and SCR, test data from the subject \#7 was not used, due to issue during the recording of this subject (sensor was partially detached from the skin).
Two different techniques were investigated to standardise the features: (i) online (standardisation parameters $\mu$ and $\sigma$ are computed on the training partition and used on all partitions), and (ii) speaker dependent ($\mu$ and $\sigma$ are computed and applied on features of each subject).
In order to compensate time reaction of the raters, a time delay $D$ is applied to the gold-standard, by shifting back in time the values of the time-series (last value was duplicated), with a delay ranging from zero to eight seconds, and a step of 400ms.

As machine learning, we used a linear Support Vector Machine (SVM) to perform the regression task with the liblinear library \cite{Fan08-LAL}; the L2-regularised L2-loss dual solver was chosen (option \textsc{-s 12}) and a unit bias was added to the feature vector (option \textsc{-B 1}), all others parameters were kept to default. The complexity of the SVM was optimised in the range [$10^{-5} - 10^0$]. In order to compensate for scaling and bias issues in the predictions, but also noise in the data, we used the same post-processing chain as employed in \cite{Trigeorgis16-AFE}. The window size $W$ and the time delay $D$ were optimised by a grid search with an early stopping strategy, \ie evaluations were stopped if no improvement was observed over the best score after two iterations. Experiments were always performed for both standardisation strategies, \ie online and per speaker. 

The best value of complexity, window size, time delay, and standardisation method were obtained by maximising the performance - measured as CCC - on the development partition with the model, learned on the training partition. Table \ref{tab:wsize} lists the best parameters for $W$ and $D$, for each modality and emotional dimension, and shows that, the valence generally requires longer window size (to extract features) and time delay (to compensate for reaction time) than for arousal; $\bar{W}_A=5.3$, $\bar{W}_V=9.3$, $\bar{D}_A=1.2$, $\bar{D}_V=1.8$. Moreover, the results show that, a separate processing of features related to ECG, \ie HRHV, and those related to EDA, \ie SCL, and SCR, is justified as the best parameters obtained those signals differ from the ones obtained on their original signal.
Regarding the standardisation technique, the online approach worked best for audio on both dimensions, and video data on valence, whereas standardisation of the features per subject worked best for all physiological features.

Mono-modal performance is reported in Table \ref{t:baseline_MASC}. Results show that, a significant improvement has been made for all modalities compared to the AVEC 2015 baseline \cite{RingevalEtAl2015_FAR}, excepted for the EDA features. In agreement with the state-of-the-art, audio features perform significantly better than any other modality on arousal, and video features on valence. Interestingly, emotion prediction from the HRHRV signal performs significantly better than with the original ECG signal, and it is ranked as the most relevant physiological descriptor for arousal, when taken alone.

Multimodal emotion recognition is performed with three different late-fusion models, because frames might be missing on the video, and EDA related features; (i) audio-ECG, used for missing video and EDA; (ii) audio-ECG-EDA, used for missing video; (iii) audio-ECG-EDA-video, used otherwise. In order to keep the complexity low, and estimate the contribution of each modality in the fusion process, we build the fusion model by a simple linear regression of the predictions obtained on the development partition, using Weka 3.7 with default parameters \cite{Hall09-TWD}. Obtained predictions were then post-processed with the same approach used for the mono-modal predictions.

\begin{equation}\label{eq:lin_reg}
   Pred_{m} = \epsilon_{m} + \sum_{i=1}^{N} \gamma_{i}*Pred_{u}(i),
\end{equation}
where $Pred_{u}(i)$ is the mono-modal prediction of the modality $i$ from the $N$ available ones (ranging from two to eight),
$\gamma_i$ and $\epsilon_m$ are regression coefficients estimated on the development partition, 
and $Pred_{m}$ is the fused prediction. 

Performance is reported in Table \ref{t:baseline_MASC}. Results show that, the baseline for the AVEC 2016 MASC is highly competitive, with the performance obtained on valence for the test partition being slightly better than the top-performer of AVEC 2015 \cite{He15-MAD}. 
%There is, however, large room for improvement of this baseline, and we are curious to see new techniques proposed by the participants, in order to push further the robustness of systems that perform automatic prediction of spontaneous and natural affective behaviours.
In order to depict the contribution of each modality in the prediction of emotion, we normalised the linear regression coefficients that were learned for the multimodal fusion model (iv) into a percentage:
\begin{equation}\label{eq:cont}
   C_i = 100* \frac{|\gamma_{i}|}{\sum_{k=1}^{N} |\gamma_{k}|},
\end{equation}
where $C_i$ is the contribution of the modality $i$ in percentage, and $\gamma_k$ are the regression coefficients of the multimodal fusion model; $N=8$.

Results show that, even if the mono-modal performance can be low for a given modality and emotion, \eg EDA for arousal or SCR for valence, cf. Table \ref{t:baseline_MASC}, 
all modalities contribute, to a certain extent, to the prediction of arousal and valence in the fusion scenario, cf. Figure \ref{f:MASC_fusion}. 
This is especially the case for SCR features on arousal and for SCL features on valence, which did not perform well when used in isolation, but contribute even outperformed appearance features in the fusion model.

\section{Conclusion }\label{s:conclusion}

\noindent We introduced AVEC 2016 -- the third combined open Audio/Visual
Emotion and Depression recognition Challenge. It comprises two sub-challenges: the detection of the affective dimensions of arousal and valence, and the estimation of a self-reported level of depression. This manuscript describes AVEC 2016's challenge conditions, data, baseline features and results. By intention, we opted to use open-source software and the highest possible transparency and realism for the baselines by refraining from feature space optimisation and optimising on test data. This should improve the reproducibility of the baseline results.

\section*{Acknowledgements}
The research leading to these results has received funding from the EC's 7th Framework Programme through the ERC Starting Grant No.\ 338164 (iHEARu), and the European Union's Horizon 2020 Programme through the Innovative Action No.\ 645094 (SEWA), and the Research Innovative Action No.\ 645378 (ARIA-VALUSPA), and No.\ 688835 (DE-ENIGMA).

\bibliographystyle{abbrv}
\bibliography{avec2016}  % sigproc.bib is the name of the Bibliography in this case
% ACM needs 'a single self-contained file'!
%\balancecolumns

\end{document}